\documentclass{article}

\usepackage[preprint]{neurips_2026}

\usepackage[utf8]{inputenc} %
\usepackage[T1]{fontenc}    %
\usepackage{hyperref}       %
\usepackage{url}            %
\usepackage{booktabs}       %
\usepackage{amsfonts}       %
\usepackage{nicefrac}       %
\usepackage{microtype}      %
\usepackage{xcolor}         %

\usepackage{xspace}
\usepackage{amsmath}
\usepackage[ruled,vlined,linesnumbered]{algorithm2e}

\usepackage{multirow}    %
\usepackage[table]{xcolor} %
\definecolor{LightGray}{gray}{0.94}
\definecolor{LightCyan}{rgb}{0.88,1,1}
\usepackage{makecell} %
\usepackage{pifont}   %

\newcommand{\cmark}{\textcolor{green!70!black}{\ding{51}}} %
\newcommand{\xmark}{\textcolor{red!70!black}{\ding{55}}}   %

\usepackage{tcolorbox}
\usepackage{enumitem}
\newtcolorbox{promptbox}[1][]{
  colback=gray!5!white,
  colframe=gray!75!black,
  title={\textbf{System Prompt}},
  fonttitle=\bfseries\sffamily,
  arc=0.5mm,
  boxrule=0.8pt,
  left=6pt, right=6pt, top=6pt, bottom=6pt,
  #1
}

\usepackage{listings}
\newcommand{\apex}{\mbox{\textsc{APEX}}\xspace}
\newcommand{\apexfull}{{\textbf{A}utomated \textbf{P}rompt \textbf{E}ngineering e\textbf{X}pert}\xspace}

\title{\apex: \apexfull \\ with Dynamic Data Selection}

\author{%
Fei Wang \\
Google \\
\small{feiwangnlp@google.com} 
\And
Si Si \\
Google \\
\small{sisidaisy@google.com}
\And
Cho-Jui Hsieh \\
UCLA \\
\small{chohsieh@cs.ucla.edu}
\And
Inderjit S. Dhillon \\ 
Google \\
\small{isd@google.com}
}

\begin{document}

\maketitle

\begin{abstract}
Large Language Models are highly sensitive to prompt formulation, necessitating automatic prompt optimization to unlock their full potential. While evolutionary algorithms have emerged as the dominant paradigm, they suffer from a critical bottleneck: data efficiency. Current methods treat the development dataset as a static benchmark, wasting significant compute budget on uninformative data. In this work, we introduce \textbf{\apex}~(\apexfull), a novel framework that optimizes the \textit{data usage} alongside the prompt search. \apex dynamically stratifies the dataset into \texttt{Easy}, \texttt{Hard}, and \texttt{Mixed} tiers based on the optimization lineage. By prioritizing the \texttt{Mixed} tier, which identifies the data where the LLM has mixed performance, we identify two high-leverage subsets: the \textbf{addressable frontier} for generating informative mutations and the \textbf{rank-sensitive frontier} for distinguishing candidate quality. We evaluate \apex across three diverse benchmarks: IFBench, SimpleQA Verified, and FACTS Grounding. Under a fixed budget of 5,000 evaluation calls, due to its data efficiency, \apex outperforms the initial prompt by an average of 11.2\% on Gemini 2.5 Flash and 6.8\% on Gemma 3 27B, demonstrating that a data-centric approach is key to efficient and effective prompt optimization. 
\end{abstract}

\section{Introduction}

Large Language Models (LLMs) have demonstrated remarkable capabilities across instruction following, fact retrieval, and complex reasoning. However, their performance remains notoriously sensitive to the prompts \citep{kojima2022large}. A minor change can often mean the difference between a correct answer and a hallucination. This fragility has given rise to automatic prompt optimization, a field dedicated to treating prompts not as static strings, but as learnable parameters that can be optimized~\citep{zhou2022large, pryzant2023automatic, deng2022rlprompt,hsieh2024automatic}.

Currently, genetic and evolutionary algorithms have emerged as the dominant paradigm in prompt optimization, exemplified by methods like APO~\citep{pryzant2023automatic} and GEPA~\citep{agrawal2025gepa}. These frameworks typically operate through an iterative two-stage process: (1) \textit{Mutation}, where the optimizer analyzes error cases from the current prompt to generate new candidate prompts, and (2) \textit{Selection}, where these candidates are evaluated and ranked by their performance to identify the better ones.

However, this paradigm faces a critical and often overlooked bottleneck: \textbf{data efficiency}. The effectiveness of both stages is heavily dependent on the quality of the data used. The mutation stage relies on identifying \textit{informative} failure cases that point towards fixable errors, while the selection stage requires a rigorous evaluation to accurately rank candidates. Current methods typically resort to naive data strategies: random sampling for mutation, and either random sampling or full-set evaluation for selection. This creates a fundamental dilemma: random sampling leads to rank instability, where superior prompts are discarded due to variance, while full-set evaluation depletes the compute budget rapidly, limiting the search to only a few generations.

We argue that the root cause of this inefficiency is the treatment of the dataset as a \textit{static benchmark}. In reality, the utility of any given data point is not fixed. It changes dynamically throughout the optimization process. As prompts evolve and improve, examples that were once informative discriminators may converge into consistently \texttt{Easy} cases~(solved by all candidates), while others persist as \texttt{Hard}~(intractable noise). Continuing to mutate based on unsolvable errors, or evaluating candidates on examples that have lost their discriminatory power, wastes a significant portion of the compute budget on signals that provide zero improvement information.

To address this, we introduce \apex~(\apexfull). APEX shifts the focus from improving the search algorithm to improving the data usage. By maintaining a lineage of prompt performance, \apex dynamically categorizes data points into \texttt{Easy}, \texttt{Hard}, and \texttt{Mixed} tiers. Our core insight is that the informative mutation and selection signals primarily lie in the \texttt{Mixed} tier, where the LLM has mixed performance on recent prompt variants. This dynamic tier constitutes two critical frontiers: the \textbf{addressable frontier} indicating volatile errors suitable for mutation, and the \textbf{rank-sensitive frontier} with ambiguous cases that differentiate candidates. By prioritizing compute to these high-leverage frontiers, \apex maximizes the optimization signal per evaluation call.

In summary, our contributions are three-fold. First, we identify the data efficiency bottleneck in current genetic prompt optimization methods and attribute it to the static treatment of dynamic data signals. Second, we propose \apex, a novel algorithm that enhances the prompt optimization process by stratifying data into dynamic tiers and targeting the ``Addressable" and ``Rank-Sensitive" frontiers. Third, we achieve performance gains across three diverse benchmarks (IFBench, SimpleQA Verified, FACTS Grounding). Under a fixed budget of 5,000 evaluation calls, \apex improves performance by 11.2\% on Gemini 2.5 Flash and 6.8\% on Gemma 3 27B, establishing a new standard for data-efficient prompt optimization.

\section{Related Work}

\subsection{Automatic Prompt Optimization}
Early research explored diverse search strategies for prompt engineering, ranging from gradient-based methods \citep{shin2020autoprompt, deng2022rlprompt,chen2024instructzero} and edit-based heuristics~\citep{prasad2023grips}, to instruction induction \citep{honovich2023instruction}. They have paved the way for LLM-driven optimization frameworks, which formalize prompt optimization by leveraging LLMs to generate candidate instructions \citep{zhou2022large, yang2023large}.

As the field matured, genetic (or evolutionary) algorithms utilizing natural language feedback emerged as the most effective paradigm. A representative method is \textbf{APO}~\citep{pryzant2023automatic}, which iteratively refines prompts by treating error feedback as a signal for mutation. Parallel work has explored other directions to improve the genetic approach, by integrating history guidance, hybrid search strategies, and carefully curated meta-prompts ~\citep{hsieh2024automatic,fernando2024promptbreeder,wangpromptagent,ye2024prompt,guo2024connecting}. Notably, AELP \citep{hsieh2024automatic} is designed for long prompts, utilizing history guidance and limiting the mutation scope to ensure stability.
Recent research has also expanded the scope of prompt optimization to agentic workflows \citep{agrawal2025gepa,opsahl2024optimizing}, in-context learning \citep{wu2024prompt, wan2024teach}, and multimodal scenarios \citep{liu2024language,wan2025maestro}.
Among them, \textbf{GEPA} \citep{agrawal2025gepa} employs a Genetic-Pareto framework to optimize agents towards a multi-objective frontier.

Despite these advancements, the field has largely overlooked the data-centric perspective of prompt optimization. Most existing methods treat the dataset as a static benchmark, ignoring the varying informational value of different examples. \apex address this fundamental inefficiency by improving the data selection process itself. By dynamically tracking the rank-sensitivity and solvability of datapoints, \apex allocates compute wisely.

\subsection{Data Selection for LLM Alignment}
The importance of data quality over quantity for LLM alignment was highlighted by the LIMA hypothesis \citep{zhou2023lima}, which provided empirical evidence that a small, curated set of examples is sufficient for alignment. This insight catalyzed a wave of automated data selection methods. AlpaGasus \citep{chen2024alpagasus} filters low-quality data to improve fine-tuning efficiency, while LESS \citep{xia2024less} selects data based on gradient similarity. Additionally, influence-based approaches \citep{choe2025what} utilize influence functions to identify which training examples most affect downstream capabilities. More recent work has shifted towards dynamic and active data curation, such as Data Advisor \citep{wang2024data}, which dynamically curates data to improve safety alignment.

Parallel to these training-time advances, data selection has recently garnered attention in prompt optimization. Notable approaches, such as those by \citet{diao2024active} and \citet{dong2025model}, employ active learning or combine data similarity and model confidence heuristics to curate examples. However, the application scope of these methods is limited by restrictive dependencies, such as human-in-the-loop annotation, reliance on internal model signals (e.g., logits), or restriction to classification tasks. These requirements render them unsuitable for general generative tasks under fully automated, black-box API settings, and non-comparable to our work.
\apex extends the principles of identifying informative data and dynamic curation to prompt optimization, offering a fully automated, black-box solution.

\begin{figure*}[!t]
    \centering
    \includegraphics[width=\textwidth]{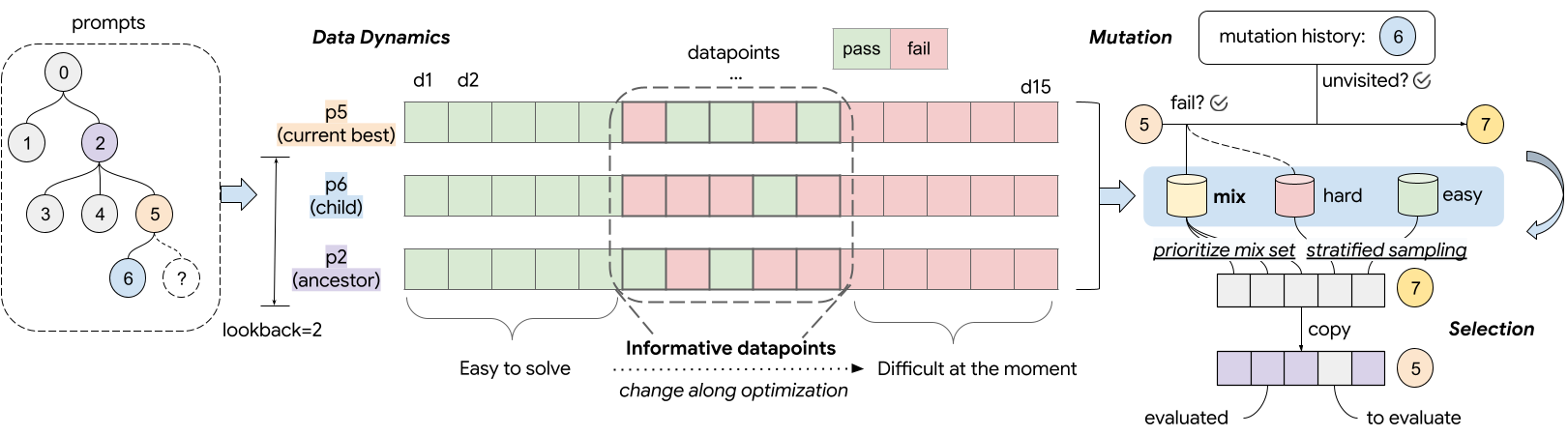} 
    \caption{\textbf{The APEX framework overview.} The optimization process begins by tracking prompt lineage (left), where historical performance across ancestors (e.g., $p_2$) and children (e.g., $p_6$) is analyzed over a defined lookback window. In the \textbf{Data Dynamics} phase (center), datapoints are dynamically categorized into tiers (\textit{Easy}, \textit{Hard}, and \textit{Mixed}) based on pass/fail trajectories. The \textit{Mixed} set identifies informative datapoints where model performance is inconsistent, signaling high potential for optimization. These tiers inform the \textbf{Mutation} and \textbf{Selection} stages (right): the former leverages mutation history to explore unvisited prompt space, while the latter employs stratified sampling to efficiently rank candidate prompts. Notably, both stages prioritize the \textit{Mixed} set to focus optimization on high-signal datapoints. Evaluation is incremental, skipping previously evaluated datapoints to maximize efficiency.}
    \label{fig:apex}
\end{figure*}

\section{Bottleneck of Prompt Optimization}

\subsection{Preliminary: Prompt Optimization}
Prompts serve as the primary interface for interacting with LLMs at the inference stage. For a given user intent, there exists an expansive space of potential prompt variants, many of which yield significantly different performance outcomes. The goal of prompt optimization is to systematically identify an optimal prompt $P^*$ that maximizes a predefined performance metric on a target LLM:
\begin{equation}
P^* = \arg\max_{P} \mathbb{E}_{x \sim \mathcal{D}^*} [f(x, \text{LLM}(P, x))],
\end{equation}
where $\mathcal{D}^*$ represents the oracle data distribution and $f(\cdot)$ is an evaluation function (e.g., string-matching accuracy or model-based grading) that scores the LLM's response relative to input $x$. 

To simplify notation across the optimization trajectory, we define $s(P, x) \in \{0, 1\}$ as the binary evaluation outcome of prompt $P$ on datapoint $x$, where only a perfect score yields a pass ($1$) and any partial credit is treated as a failure ($0$).

In practice, both the LLM and the evaluation function $f$ are treated as black boxes where gradients are unavailable. Consequently, derivative-free genetic algorithms, such as APO \citep{pryzant2023automatic} and GEPA \citep{agrawal2025gepa}, have emerged as leading methods. These methods typically iterate through two primary stages on a collected dataset $\mathcal{D}$:

\textbf{Mutation:} To explore the prompt space, a meta-optimizer $\text{LLM}_{\text{meta}}$ (typically the target LLM itself) generates a new prompt $P_{\text{new}}$ conditioned on an existing prompt $P_{\text{curr}}$ and a batch of its error cases $\mathcal{E}_{\text{curr}} \subset \{x \in \mathcal{D} \mid s(P_{\text{curr}}, x) = 0\}$:
\begin{equation}
P_{\text{new}} \sim \text{LLM}_{\text{meta}}(P_{\text{curr}}, \mathcal{E}_{\text{curr}}).
\end{equation}

\textbf{Selection:}
The new candidate prompt $P_{\text{new}}$ is evaluated alongside the current population $\mathcal{P}_t$. The subsequent generation $\mathcal{P}_{t+1}$ is constructed by selecting the $k$ highest-performing prompts based on their aggregate scores on $\mathcal{D}$:
\begin{equation}
\mathcal{P}_{t+1} = \text{Top-}k \left( \mathcal{P}_t \cup \{P_{\text{new}}\} \mid \mathcal{D} \right).
\end{equation}

\subsection{Problem: Data Efficiency}
\label{sec:problem}

Despite their success, data efficiency remains a critical but overlooked bottleneck in both the mutation and selection stages, hindering the scalability of existing frameworks. We formalize the global optimization state as a \textbf{sparse score matrix} $\mathbf{S} \in \{0, 1, \emptyset\}^{|\mathcal{P}| \times |\mathcal{D}|}$, where each entry $S_{j, i}$ represents the performance of prompt $P_j$ on datapoint $x_i \in \mathcal{D}$:
\begin{equation}
S_{j, i} = s(P_j, x_i) \in \{0, 1\}, \quad (\text{or } \emptyset \text{ if not applicable}).
\end{equation}

\paragraph{Inefficiency in Mutation.}
The mutation distribution produced by $\text{LLM}_{\text{meta}}$ highly depends on $\mathcal{E}_{\text{curr}}$. Current methods typically rely on random sampling of error cases, which fails to account for the \textbf{optimization trajectory}, rendering the mutation process a source of instability rather than monotonic improvement.

Optimization performance is often governed by a hierarchy of ``fixability,'' where certain errors serve as logical precursors to others. For instance, if error samples' batch $\mathcal{E}_b$ contains errors only addressable after foundational issues in $\mathcal{E}_a$ are resolved, the trajectory must follow the sequence $\mathcal{E}_a \rightarrow \mathcal{E}_b$. Reversing this sequence or sampling randomly leads to high-variance updates and ``forgetting,'' as the $\text{LLM}_{\text{meta}}$ attempts to solve high-level failures without the necessary prompt refinements established in earlier stages.

This lack of trajectory awareness also frequently precipitates the trap where $\mathcal{E}_{\text{curr}}$ becomes dominated by ``impossible'' cases that are fundamentally beyond the capacity of the target LLM. Attempting to optimize for these points misleads the $\text{LLM}_{\text{meta}}$, stalling progress. Notably, the optimal data for mutation is non-static; it changes dynamically along the optimization process as different error types become addressable at different stages of prompt maturity. 

\paragraph{Inefficiency in Selection.}
The selection stage represents the primary computational overhead of the optimization loop, often accounting for more than 90\% of total costs.\footnote{Generating a mutation often requires fewer than three API calls, whereas the selection phase requires evaluating the new prompt across the entire dataset or multiple candidates on subsets, frequently requiring hundreds of API calls, as reported in APO and GEPA.} This creates a fundamental trade-off between selection accuracy and the number of iterations $T$ possible within a fixed budget. While comprehensive evaluation ensures precise ranking, it is computationally prohibitive. Conversely, stochastically sampling a subset introduces estimation noise that leads to rank inversion, where suboptimal prompts are mistakenly retained. 

Crucially, absolute performance estimates are often redundant. For selection, the priority is the relative rank of candidates. Because the population consists of parents, children, and siblings that share significant structural or semantic similarities, their performance is identical for a large portion of $x \in \mathcal{D}$. The true bottleneck is the failure to isolate \textbf{discriminative data}:
\begin{equation} \mathcal{D}_{\text{disc}} = \{x_i \in \mathcal{D} \mid \exists P_a, P_b \in \mathcal{P}_t : s(P_a, x_i) \neq s(P_b, x_i) \}. \end{equation}
Existing frameworks waste their budget evaluating $x \notin \mathcal{D}_{\text{disc}}$, where all candidates perform identically. Evaluating these points provides zero information for selection, effectively reducing the number of meaningful updates possible within a fixed compute budget.

\section{Method: \apex}

The \apex framework (Figure \ref{fig:apex}) addresses the core bottlenecks of data inefficiency by dynamically re-evaluating the utility of each datapoint in the optimization loop (Algorithm \ref{alg:apex}). Instead of treating the development set as a static benchmark, \apex categorizes data based on its historical interaction with the prompt lineage to guide both mutation and selection.

\begin{figure}[t]
    \centering
    
    \begin{minipage}{0.55\textwidth}
        \begin{algorithm}[H]
            \setcounter{AlgoLine}{0}
            \DontPrintSemicolon
            \small
            \SetKwInOut{Input}{Input}\SetKwInOut{Output}{Output}
            \caption{APEX (Mutation \& Selection)}
            \label{alg:apex}
            
            \Input{Initial $P_0$, Iterations $T$, Budget $N$, Batch $m$, Init anchor $\alpha_0$, Lookback $k$}
            \Output{Optimized Prompt $P^*$}
            
            $P_{\text{curr}} \leftarrow P_0; \quad \alpha \leftarrow \alpha_0$\;
            $\mathcal{H} \leftarrow \{ \text{EvaluateFull}(P_0, \mathcal{D}) \}$\;
            
            \For{$t = 1 \dots T$}{
                \tcp{1. Dynamic Data Stratification}
                $\mathcal{B} \leftarrow \text{Stratify}(\mathcal{D}, \mathcal{H}, k)$ \tcp*{Alg. \ref{alg:stratify}}
                
                \vspace{2pt}
                \tcp{2. Trajectory-Guided Mutation}
                $\mathcal{E} \leftarrow \text{Sample } m \text{ unused from } \mathcal{B}_{M, 0} \cup \mathcal{B}_{H, 0}$\;
                $C \leftarrow \text{LLM}_{\text{meta}}(\text{Critique}(P_{\text{curr}}, \mathcal{E}))$\;
                $P_{\text{new}} \leftarrow \text{LLM}_{\text{meta}}(\text{Mutate}(P_{\text{curr}}, C))$\;
                
                \vspace{2pt}
                \tcp{3. Rank-Sensitive Selection}
                $\mathcal{D}_{\text{req}} \leftarrow \mathcal{B}_{M, \emptyset}$\;
                $R \leftarrow N - |\mathcal{D}_{\text{req}}|$ \tcp*{Remaining budget}
                $\rho_{\text{mix}} \leftarrow \text{PassRate}(\mathcal{B}_{M}); \;\; \rho_{\text{all}} \leftarrow \text{PassRate}(\mathcal{D})$\;
                $k_{\text{pos}} \leftarrow \lfloor \min(\alpha, \rho_{\text{mix}}, \rho_{\text{all}}) \cdot R \rfloor; \;\; k_{\text{neg}} \leftarrow R - k_{\text{pos}}$\;
                
                $\mathcal{D}_{\text{pos}} \leftarrow \text{Sample } k_{\text{pos}} \text{ prioritizing } \mathcal{B}_{M, 1} \text{ then } \mathcal{B}_{E, \emptyset}$\;
                $\mathcal{D}_{\text{neg}} \leftarrow \text{Sample } k_{\text{neg}} \text{ prioritizing } \mathcal{B}_{M, 0} \text{ then } \mathcal{B}_{H, \emptyset}$\;
                
                $\mathcal{D}_{\text{eval}} \leftarrow \mathcal{D}_{\text{req}} \cup \mathcal{D}_{\text{pos}} \cup \mathcal{D}_{\text{neg}}$\;
                Evaluate $P_{\text{new}}$ and $P_{\text{curr}}$ on $\mathcal{D}_{\text{eval}}$\;
                
                \vspace{2pt}
                \If{$P_{\text{new}}$ is better than $P_{\text{curr}} \text{ on } \mathcal{D}_{\text{eval}}$}{
                    $P_{\text{curr}} \leftarrow P_{\text{new}}$\;
                    $\alpha \leftarrow \alpha + \beta$ \tcp*{Anneal anchor}
                }
                Update $\mathcal{H}$ with Evaluation of $P_{\text{new}}$ and $P_{\text{curr}}$ on $\mathcal{D}_{\text{eval}}$\;
            }
            \Return{$P_{\text{curr}}$}\;
        \end{algorithm}
    \end{minipage}\hfill
    \begin{minipage}{0.42\textwidth}
        \begin{algorithm}[H]
            \setcounter{AlgoLine}{0}
            \DontPrintSemicolon
            \small
            \SetKwInOut{Input}{Input}\SetKwInOut{Output}{Output}
            \caption{Dynamic Data Stratification}
            \label{alg:stratify}
            
            \Input{Dataset $\mathcal{D}$, History $\mathcal{H}$, Lookback $k$}
            \Output{Granular Data Buckets $\mathcal{B}_{t, s}$}
            
            Initialize empty buckets $\mathcal{B}_{t, s}$\;
            
            \For{each $x_i \in \mathcal{D}$}{
                \tcp{Retrieve the last k non-empty outcomes}
                $\mathcal{R}_i \leftarrow \text{Last } k \text{ valid scores } s(P, x_i) \text{ from } \mathcal{H}$\;
                
                \vspace{4pt}
                \tcp{Determine semantic tier}
                \If{$\text{Set}(\mathcal{R}_i) \equiv \{1\}$}{
                    $t \leftarrow \text{E (Easy)}$\;
                }
                \ElseIf{$\text{Set}(\mathcal{R}_i) \equiv \{0\}$}{
                    $t \leftarrow \text{H (Hard)}$\;
                }
                \Else{
                    $t \leftarrow \text{M (Mixed)}$\;
                }
                
                \vspace{4pt}
                \tcp{Lookup cached status (yields $\emptyset$ if uneval)}
                $s \leftarrow \text{Retrieve } s(P_{\text{curr}}, x_i) \text{ from } \mathcal{H}$\;
                $\mathcal{B}_{t, s} \leftarrow \mathcal{B}_{t, s} \cup \{x_i\}$\;
            }
            \vspace{4pt}
            \Return{$\mathcal{B}$}\;
        \end{algorithm}
    \end{minipage}

\end{figure}

\subsection{Data Dynamics for Prompt Optimization}
The core philosophy of \apex is to move beyond static data by characterizing how individual datapoints influence the optimization trajectory. We categorize the interaction between the prompt lineage and the dataset into three distinct dynamics: \texttt{Easy}, \texttt{Hard}, and \texttt{Mixed} (Algorithm \ref{alg:stratify}).

To accurately capture these dynamics without overfitting to stale data, we treat the evaluation history $\mathcal{H}$ as an ordered sequence of prompts. For any datapoint $x_i \in \mathcal{D}$, let $\mathcal{H}^{(i)}_{\text{valid}}$ be the sub-sequence of historical prompts that have actively evaluated $x_i$ (i.e., $s(P, x_i) \neq \emptyset$). We define its local history $\mathcal{R}_i$ as the set of outcomes from the $k$ most recent prompts in $\mathcal{H}^{(i)}_{\text{valid}}$:
\begin{equation}
    \mathcal{R}_i = \left\{ s(P, x_i) \mid P \in \text{last}_k\big(\mathcal{H}^{(i)}_{\text{valid}}\big) \right\}.
\end{equation}
The lookback window $k$ acts as a dynamic slice on the sparse score matrix $\mathbf{S}$, ensuring the optimizer always has a consistent sample size of relevant, recent behavior rather than outdated signals.

We partition the dataset $\mathcal{D}$ into three semantic tiers based on the variance of $\mathcal{R}_i$:
\begin{equation}
    \text{Tier}(i) = 
    \begin{cases} 
      \text{E (Easy)} & \text{if } \text{Set}(\mathcal{R}_i) \equiv \{1\} \\
      \text{H (Hard)} & \text{if } \text{Set}(\mathcal{R}_i) \equiv \{0\} \\
      \text{M (Mixed)} & \text{if } \text{Set}(\mathcal{R}_i) \equiv \{0, 1\}
    \end{cases}
\end{equation}
\begin{itemize}[leftmargin=*, topsep=0pt]
    \item \textbf{Easy:} Points consistently solved by the lineage. Re-evaluating them provides minimal signal.
    \item \textbf{Hard:} Points consistently failed. These represent data that are currently intractable.
    \item \textbf{Mixed:} Points exhibiting volatility. These represent the rank-sensitive frontier for evaluation and the most probable targets for improvement.
\end{itemize}

By intersecting these historical tiers $T \in \{E, H, M \}$ with the evaluation outcome $s \in \{1, 0, \emptyset\}$ under the \textit{current} prompt $P_{\text{curr}}$, we partition the dataset $\mathcal{D}$ into nine disjoint subsets, denoted as $\mathcal{B}_{T, s}$. This notation provides a granular view of the data state. For example, $\mathcal{B}_{M, 0} \subseteq \mathcal{D}$ represents the subset of historically Mixed instances that are currently failing, while $\mathcal{B}_{E, \emptyset}$ represents Easy instances that were skipped in the current pass.

\begin{table}[t]
\caption{\textbf{Main Results.} Accuracy (\%) {\scriptsize{$\pm$ Std Dev}} and absolute gain ({\color{blue}$\Delta$}) over the initial prompt. APEX consistently outperforms baselines.}
\label{tab:main-results}
\small
\centering
\begin{tabular}{l c c c c}
\toprule
\textbf{Method} & \textbf{IFBench} & \textbf{SimpleQA Verified} & \textbf{FACTS Grounding} & \textbf{Average} \\
\midrule
\multicolumn{5}{c}{\textit{Gemini 2.5 Flash}} \\
\midrule
\cellcolor{LightGray}Initial Prompt & \cellcolor{LightGray}38.5 {\scriptsize $\pm$ 1.2} {\color{gray}(---)} & \cellcolor{LightGray}23.6 {\scriptsize $\pm$ 0.6} {\color{gray}(---)} & \cellcolor{LightGray}85.8 {\scriptsize $\pm$ 1.1} {\color{gray}(---)} & \cellcolor{LightGray}49.3 {\color{gray}(---)} \\
GEPA & 41.2 {\scriptsize $\pm$ 0.9} {\color{blue}(+2.7)} & 28.8 {\scriptsize $\pm$ 0.8} {\color{blue}(+5.2)} & 93.5 {\scriptsize $\pm$ 0.0} {\color{blue}(+7.7)} & 54.5 {\color{blue}(+5.2)} \\
APO ($|D|$) & 43.7 {\scriptsize $\pm$ 1.2} {\color{blue}(+5.2)} & 27.4 {\scriptsize $\pm$ 2.6} {\color{blue}(+3.8)} & 89.7 {\scriptsize $\pm$ 1.9} {\color{blue}(+3.9)} & 53.6 {\color{blue}(+4.3)} \\
APO ($|D|/2$) & 43.5 {\scriptsize $\pm$ 1.9} {\color{blue}(+5.0)} & 25.0 {\scriptsize $\pm$ 3.5} {\color{blue}(+1.4)} & 89.9 {\scriptsize $\pm$ 0.1} {\color{blue}(+4.1)} & 52.8 {\color{blue}(+3.5)} \\
\cellcolor{LightCyan}\textbf{APEX (Ours)} & \cellcolor{LightCyan}\textbf{52.3 {\scriptsize $\pm$ 1.4} {\color{blue}(+13.8)}} & \cellcolor{LightCyan}\textbf{35.0 {\scriptsize $\pm$ 2.3} {\color{blue}(+11.4)}} & \cellcolor{LightCyan}\textbf{94.1 {\scriptsize $\pm$ 0.4} {\color{blue}(+8.3)}} & \cellcolor{LightCyan}\textbf{60.5 {\color{blue}(+11.2)}} \\
\midrule
\multicolumn{5}{c}{\textit{Gemma 3 27B}} \\
\midrule
\cellcolor{LightGray}Initial Prompt & \cellcolor{LightGray}33.4 {\scriptsize $\pm$ 0.7} {\color{gray}(---)} & \cellcolor{LightGray}9.4 {\scriptsize $\pm$ 0.5} {\color{gray}(---)} & \cellcolor{LightGray}80.7 {\scriptsize $\pm$ 1.6} {\color{gray}(---)} & \cellcolor{LightGray}41.2 {\color{gray}(---)} \\
GEPA & 34.1 {\scriptsize $\pm$ 0.5} {\color{blue}(+0.7)} & 9.4 {\scriptsize $\pm$ 0.5} {\color{blue}(+0.0)} & 91.7 {\scriptsize $\pm$ 0.5} {\color{blue}(+11.0)} & 45.1 {\color{blue}(+3.9)} \\
APO ($|D|$) & 35.7 {\scriptsize $\pm$ 2.4} {\color{blue}(+2.3)} & \textbf{11.5 {\scriptsize $\pm$ 0.1} {\color{blue}(+2.1)}} & 88.5 {\scriptsize $\pm$ 0.7} {\color{blue}(+7.8)} & 45.2 {\color{blue}(+4.0)} \\
APO ($|D|/2$) & 36.1 {\scriptsize $\pm$ 0.5} {\color{blue}(+2.7)} & 8.4 {\scriptsize $\pm$ 1.4} {\color{red}(-1.0)} & 85.7 {\scriptsize $\pm$ 0.9} {\color{blue}(+5.0)} & 43.4 {\color{blue}(+2.2)} \\
\cellcolor{LightCyan}\textbf{APEX (Ours)} & \cellcolor{LightCyan}\textbf{39.3 {\scriptsize $\pm$ 2.0} {\color{blue}(+5.9)}} & \cellcolor{LightCyan}\textbf{11.5 {\scriptsize $\pm$ 1.2} {\color{blue}(+2.1)}} & \cellcolor{LightCyan}\textbf{93.3 {\scriptsize $\pm$ 0.1} {\color{blue}(+12.6)}} & \cellcolor{LightCyan}\textbf{48.0 {\color{blue}(+6.8)}} \\
\bottomrule
\end{tabular}
\end{table}

\subsection{Trajectory-Guided Mutation}
To resolve the stochastic instability of mutation, \apex enforces a trajectory-guided strategy. Standard optimization methods uniformly sample errors from the entire failure set, frequently retrieving ``hard'' cases ($\mathcal{B}_{H,0}$) that the current prompt logic cannot handle, leading to hallucinated corrections. 

In contrast, \apex targets the \textbf{addressable frontier} by constructing the mutation error batch primarily from the \texttt{Mixed-Fail} bucket ($\mathcal{B}_{M, 0}$). These ``soft failures'' are datapoints where the model has demonstrated capacity in the recent lineage but regressed under the current variation. Correcting these regressions stabilizes the trajectory. 

Furthermore, to ensure coverage maximization, we maintain a usage history $\mathcal{U}$ to prevent the optimizer from repeatedly overfitting to the same recurring failures. Errors for mutation are exclusively drawn from unvisited failures:
\begin{equation}
    e \in \left\{ x_i \mid x_i \in (\mathcal{B}_{M,0} \cup \mathcal{B}_{H,0}), \, x_i \notin \mathcal{U} \right\}.
\end{equation}
Once this pool is exhausted, the usage history is reset, guaranteeing broad exploration of the error surface.

\subsection{Rank-Sensitive Evaluation}
The selection stage reduces evaluation cost by focusing on the \textbf{rank-sensitive subset}, which contains datapoints where the new prompt is most likely to diverge from its parent.
We construct evaluation dataset $\mathcal{D}_{\text{eval}}$ using a tiered sampling strategy to fill the budget $N$. First, we identify volatile unknowns, data points that are historically mixed but were skipped during the previous evaluation. These provide the highest information gain and form the required baseline:
$\mathcal{D}_{\text{req}} = \mathcal{B}_{M, \emptyset}.$
To fill the remaining evaluation budget $R = N - |\mathcal{D}_{\text{req}}|$, we employ stratified sampling to balance stability (anchors) and error correction. We sample a positive set $\mathcal{D}_{\text{pos}}$ (prioritizing $\mathcal{B}_{M,1}$ to catch regressions, followed by $\mathcal{B}_{E,\emptyset}$) and a negative set $\mathcal{D}_{\text{neg}}$ (prioritizing $\mathcal{B}_{M,0}$ to confirm fixes). 

The size of these sets is governed by an \textbf{anchor ratio} $\alpha_t$, which dictates the proportion of the budget dedicated to verifying positive stability. This ratio follows an annealing schedule, growing when a prompt update is successful:
\begin{equation}
    \alpha_{t+1} = \alpha_t + \beta \cdot \mathbb{I}(P_{\text{new}} \succ P_{\text{curr}}).
\end{equation}
As the prompt improves, $\alpha$ increases, effectively ``locking in'' mastered logic by dedicating more evaluation budget to preventing regressions. To ensure the anchor size is calibrated to the model's actual competence, we clamp $\alpha_t$ against the pass ratios of the mixed tier ($\rho_{\text{mix}}$) and the global dataset ($\rho_{\text{all}}$):
\begin{equation}
    k_{\text{pos}} = \lfloor \min(\alpha_t, \rho_{\text{mix}}, \rho_{\text{all}}) \cdot R \rfloor, \quad k_{\text{neg}} = R - k_{\text{pos}}.
\end{equation}

Finally, to minimize computational overhead, \apex employs an \textbf{incremental evaluation} scheme. If an outcome $s(P_{\text{curr}}, x_i)$ for a sampled point already exists in the history $\mathcal{H}$, we retrieve it directly from memory, executing full inference only for the new candidate.

\section{Experiments}

\begin{figure}[t]
    \centering
    
    \begin{minipage}{0.48\textwidth}
        \centering
        \includegraphics[width=\linewidth]{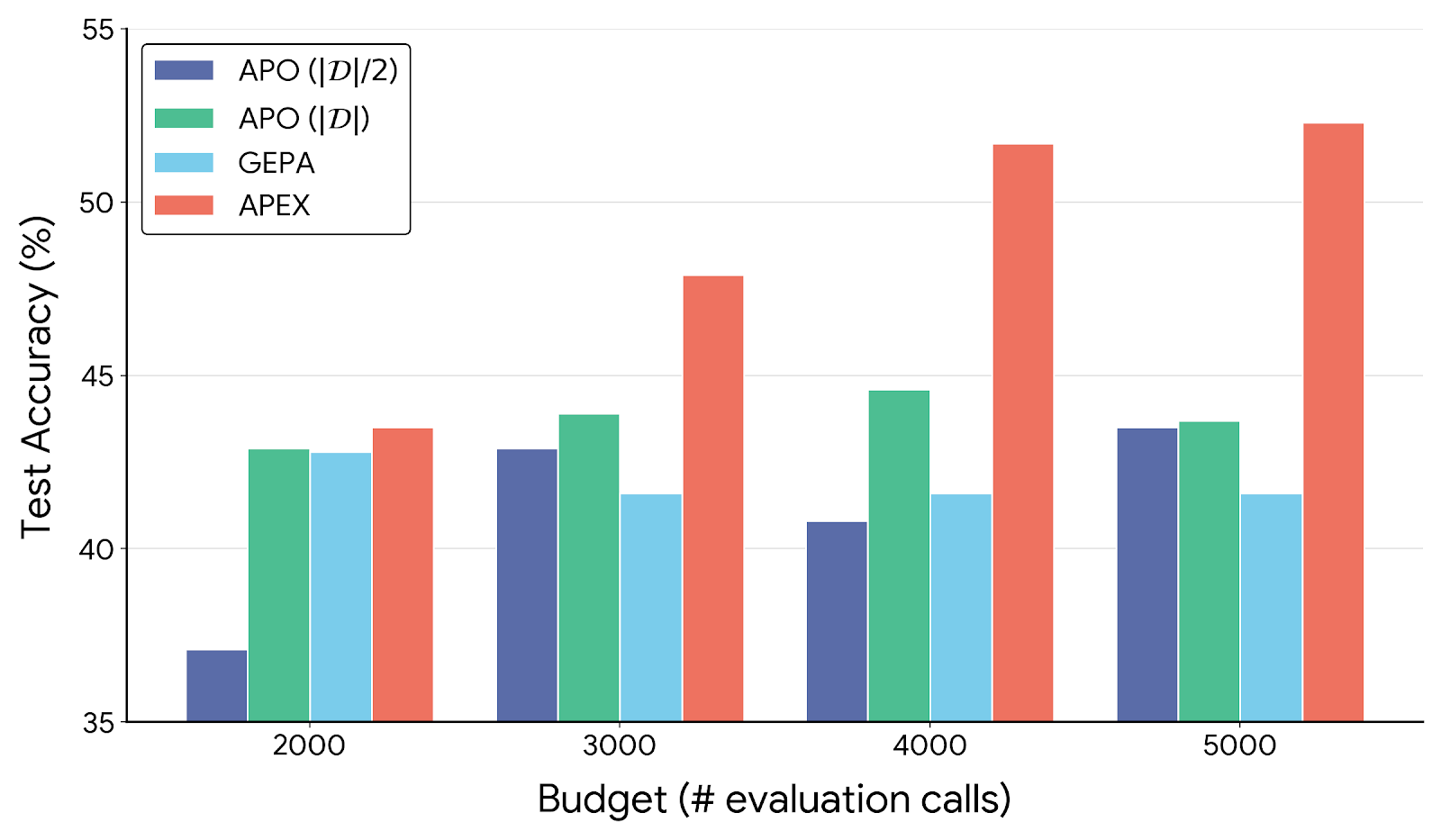}
        \caption{\textbf{Comparison of test accuracy versus budget} (i.e., number of evaluation calls) on IFBench with Gemini 2.5 Flash. The performance margin between APEX and baselines becomes larger as budget increases.}
        \label{fig:curve}
    \end{minipage}\hfill %
    \begin{minipage}{0.48\textwidth}
        \centering
        \includegraphics[width=\linewidth]{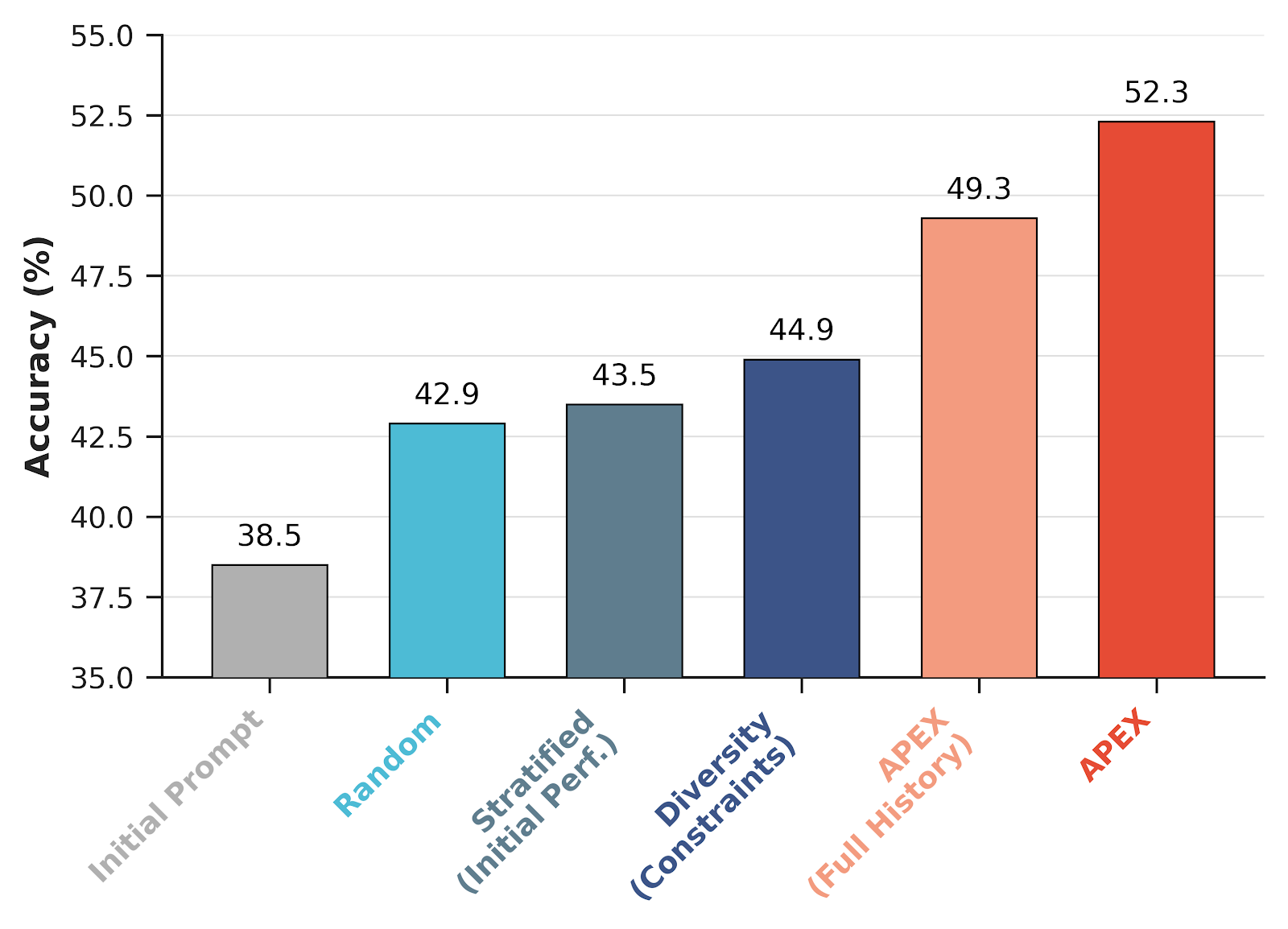}
        \caption{\textbf{Ablation study of data selection strategies}. Comparison of test accuracy across different selection methods using Gemini-2.5-Flash on IFBench.}
        \label{fig:ablation}
    \end{minipage}
    
\end{figure}

\subsection{Experimental Setting}

\paragraph{General Setup.}
We employ \textbf{Gemini-2.5-Flash} \citep{comanici2025gemini} and \textbf{Gemma-3-27B} \citep{team2025gemma} as our underlying LLMs. For each experiment, we utilize the same model for both the meta-optimizer ($\text{LLM}_{meta}$) and the target model ($\text{LLM}$). 
To enforce a fair comparison across methods with varying sample efficiency, we restrict all optimizers to a fixed global budget of 5,000 calls to the evaluation function.

\paragraph{Datasets.}
We evaluate \apex on three distinct tasks representing instruction following, parametric factuality, and grounded generation.
\textbf{IFBench} \citep{pyatkin2025generalizing}: A benchmark for precise instruction following on verifiable constraints. The evaluation function verify adherence to constraints using code-based verifiers. Following the protocol of \citet{agrawal2025gepa}, we sample 500 data points from the IF-RLVR split for optimization and reserve 294 examples from IFBench for testing. Notably, the constraints in the development set and the test set have no overlap. 
\textbf{SimpleQA Verified} \citep{haas2025simpleqa}: Assesses short-form factuality based on the model's parametric knowledge across diverse topics. The evaluation compares the model's prediction against the gold-standard answer. We randomly partition the data into 700 examples for optimization and 300 for testing. 
\textbf{FACTS Grounding} \citep{jacovi2025facts}: Evaluates the model's ability to generate long-form responses with respect to a provided context document. The evaluation assesses both response quality and grounding quality using a multi-phase LLM-based judge system. We randomly partition the data into 660 examples for optimization and 200 for testing.

\paragraph{Baselines.}
We compare \apex against the following leading prompt optimization frameworks.
\textbf{APO} \citep{pryzant2023automatic}: A genetic based prompt optimization method that remains the best-performing after its release \citep{wan2024teach}.\footnote{\url{https://github.com/microsoft/LMOps/tree/main/prompt_optimization}} It utilizes reflective mutation based on textual feedback and employs a sampling-based selection mechanism with a controllable budget per iteration. 
Since APO allows for a controllable evaluation budget, we include two baselines with per-iteration budgets set to the size of the development set ($|D|$) and half that size ($|D|/2$).
\textbf{GEPA} \citep{agrawal2025gepa}: A genetic framework designed for agentic workflows.\footnote{\url{https://github.com/gepa-ai/gepa}} It evolves prompts via reflective mutation and employs a Pareto-based selection strategy on the full development set. 
\paragraph{Implementation Details.}
For \apex, we employ a lineage lookback window of $k=5$ to construct the history for data categorization. The mutation batch size is set to $m=5$. For the rank-sensitive evaluation, we set the evaluation budget per iteration to $T=100$. The anchor ratio initializes at $\alpha = 0.2$ and increments by $\beta = 0.03$ when a better prompt is found, gradually stabilizing the evaluation set as the prompt converges.

\subsection{Main Results}
Table \ref{tab:main-results} presents the comparative performance of \apex against baseline methods. \apex consistently outperforms baselines across both LLMs and all three benchmarks, achieving the highest average accuracy. On Gemini 2.5 Flash, the method secures a mean accuracy of 60.5\%~(+11.2\% over the initial prompt), obviously surpassing the strongest baseline, GEPA, which reaches 54.5\%. This observation is consistent on Gemma 3 27B, where \apex maintains its lead. Notably, this performance gap is particularly pronounced on complex instruction-following tasks and when optimizing stronger LLMs, highlighting the method's scalability.
Broadly, these results validate that the bottleneck in prompt optimization is not merely the search algorithm, but the \textit{effective optimization signals}. While baselines struggle with the trade-off between signal cost and signal noise, \apex demonstrates that actively shaping the data distribution allows for directed and stable improvement.

A deeper analysis of the experimental results isolates two distinct failure modes in existing prompt selection strategies. First, methods reliant on full-set evaluation (GEPA) suffer from \textit{limited search depth}. While the evaluation signal is reliable, the prohibitive computational cost restricts the depth of optimization trajectories. Second, random subsampling strategies (APO) lack \textit{evaluation stability}. The reliance on small random sets introduces high variance, failing to reliably distinguish between superior and inferior prompts. This instability is evident in that aggressive subsampling with fewer evaluation samples to exchange for more iterations (APO $|D|/2$) may lead to performance regression compared to the conservative baseline (APO $|D|$).
Furthermore, these evaluation issues exacerbate the inherent limitations of random mutation. GEPA truncates the search before the process can stabilize, while APO exhibits performance regression by corrupting the selection signal with noise. Consequently, baselines fail to achieve \textit{stable and monotonic improvement}. \apex effectively navigates these trade-offs through its dynamic data selection strategy. By prioritizing the \textit{Addressable Frontier} for mutation and the \textit{Rank-Sensitive Frontier} for selection, the method achieves better performance across multiple scenarios.

Beyond aggregate metrics, we observe that the magnitude of improvement is strongly correlated with the interplay between task characteristics and LLM capability. Prompt optimization yields the most significant returns in a ``promising region" where the model possesses latent knowledge but lacks the instruction alignment to express it. This is evident in the contrast between easy tasks like FACTS Grounding, where strong base capabilities allow for substantial refinement, and difficult scenarios like Gemma 3 27B on SimpleQA Verified, where fundamental knowledge deficits impose a hard ceiling on improvements regardless of the optimization strategy.

\begin{table}[t]
    \centering
    \begin{minipage}{0.48\textwidth}
        \centering
        \caption{Independent contributions of \apex's mutation and selection components on IFBench with Gemini-2.5-Flash.}
        \label{tab:apex_ablation}
         \begin{tabular}{ccc}
            \toprule
            \textbf{\small \makecell{Trajectory-Guided \\ Mutation}} & \textbf{\small \makecell{Rank-Sensitive \\ Selection}} & \textbf{Score} \\
            \midrule
            \cmark & \cmark & 52.3 \\
            \xmark & \cmark & 50.2 \\
            \cmark & \xmark & 48.3 \\
            \xmark & \xmark & 42.9 \\
            \bottomrule
        \end{tabular}
    \end{minipage}\hfill
    \begin{minipage}{0.48\textwidth}
        \centering
        \caption{Impact of sampling from different data tiers on IFBench with Gemini-2.5-Flash. }
        \label{tab:mixed_data_ablation}
        \begin{tabular}{lc}
            \toprule
            \textbf{Method} & \textbf{Score} \\
            \midrule
            \apex & 52.3 \\
            \quad -- Random on hard and mixed & 47.3 \\
            \quad -- Random on hard & 30.3 \\
            \quad -- Random on all data & 42.9 \\
            \bottomrule
        \end{tabular}
    \end{minipage}
\end{table}

\subsection{Analysis}
To provide an in-depth understanding of the prompt optimization methods, we conduct a fine-grained analysis on IFBench with Gemini 2.5 Flash.

\paragraph{Performance across Budgets.}
Figure \ref{fig:curve} illustrates the test accuracy achieved across different evaluation budgets. GEPA exhibits competitive early performance but makes little progress thereafter, a result of \textit{insufficient exploration} constrained by high per-iteration costs. Conversely, the aggressive subsampling of APO ($|\mathcal{D}|/2$) results in \textit{instability}. In contrast, \apex maintains a \textit{sustained improvement} that becomes increasingly pronounced as the budget scales, validating the critical importance of \textit{data efficiency}.

\paragraph{Impact of Data Selection Strategy.}
Figure \ref{fig:ablation} dissects the contribution of different data selection mechanisms to the final performance. The results highlight a clear hierarchy of efficacy. \textit{Random Selection} provides a baseline improvement over the initial prompt. \textit{Stratified Sampling} balances the dataset based on \textit{initial prompt performance}, marginally improving upon random selection. Explicitly enforcing \textit{Diversity} based on constraint types provides further gains, as it introduces auxiliary guidance information. However, these methods still fall significantly short of \apex variants. The comparison between \textit{\apex (Full History)} and the complete \apex validates the hypothesis of our dynamic frontier strategy, demonstrating that removing \textit{outdated signals} to target the dynamic frontiers delivers a substantial performance margin.

\paragraph{Ablation on Mutation and Selection Methods.}
We conduct an ablation study to isolate and quantify the individual contributions of \apex's prompt mutation and selection mechanisms. As detailed in Table~\ref{tab:apex_ablation}, replacing both components with random baselines results in a severe performance drop. Introducing \apex's selection strategy independently yields a greater improvement than applying the mutation strategy alone. Crucially, the combination of both components achieves the peak overall score, highlighting a strong synergistic effect between trajectory-guided mutation and rank-sensitive selection.

\paragraph{Necessity of Prioritizing Mixed Data.}
Table~\ref{tab:mixed_data_ablation} highlights the critical role of mixed data during optimization. Sampling exclusively from the hard tier severely degrades performance, causing the prompt to overfit to narrow edge cases at the expense of general accuracy. In contrast, incorporating mixed data alongside hard examples substantially outperforms uniform random sampling across all data. This confirms that prioritizing mixed-tier examples provides a vital grounding signal, preventing catastrophic overfitting and ensuring the prompt maintains broad generalization.

\begin{figure*}[t]
    \centering
    \includegraphics[width=\linewidth]{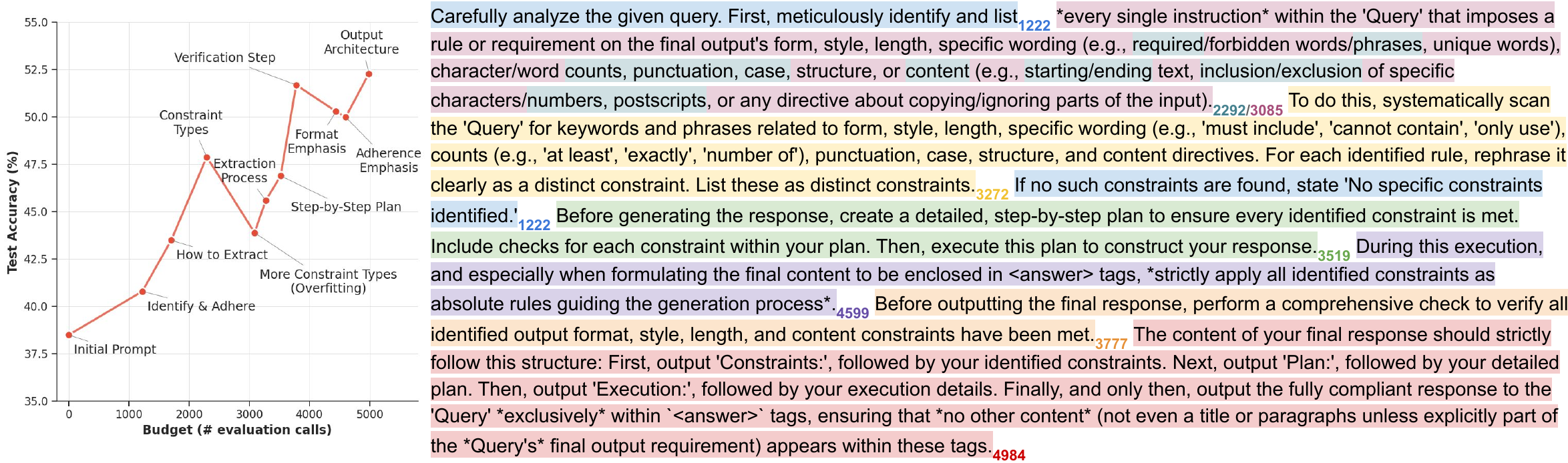}
    \caption{\textbf{Qualitative analysis of \apex prompt evolution.} The \textbf{left panel} tracks the optimization trajectory, highlighting when key instructional strategies were discovered. The \textbf{right panel} shows the final prompt, color-coded to match these milestones. Together, they demonstrate how \apex iteratively builds high-performance prompts by accumulating improvements over time.}
    \label{fig:qualitative}
\end{figure*}

\paragraph{Qualitative Example.}
The evolution of the \apex prompt in Figure \ref{fig:qualitative} demonstrates a clear shift from simple instruction-giving to establishing a rigid meta-cognitive scaffolding for the model. Early iterations focus on \textit{input analysis}, explicitly teaching the model what a constraint is.
The middle phase introduces \textit{process control}, forcing a ``plan and verify" mechanism that separates the reasoning process from the final generation. 
The final and most critical optimization creates a strict \textit{output architecture}. By mandating a specific response structure~(\texttt{Constraints} $\rightarrow$ \texttt{Plan} $\rightarrow$ \texttt{Execution} $\rightarrow$ \texttt{Answer}), the prompt acts as a forcing function.

\section{Conclusion}

In this work, we identified the static treatment of data as the fundamental efficiency bottleneck in automated prompt optimization. Existing evolutionary methods waste improved model capabilities on fixed benchmarks, treating informative and uninformative data equally. We addressed this by proposing \apex, a data-centric framework that transforms the development set from a static ruler into a dynamic pool.
By stratifying data into dynamic tiers, \apex focuses the optimization budget on the addressable frontier for prompt mutation and the rank-sensitive frontier for prompt selection.
Our empirical results across instruction following, factuality, and grounding tasks demonstrate that \apex achieves better performance under the same compute budget. 
Beyond prompt engineering, these findings suggest a broader principle for black-box optimization. As candidates evolve, the data used to optimize them must evolve together. We expect \apex to inspire future work to further explore the intersection of dynamic data curation and agent optimization.

\bibliographystyle{plainnat} %
\bibliography{reference}          %

\newpage

\appendix

\section{Analysis}
\paragraph{Impact of the Lookback Window.}
A sensitivity analysis demonstrates \apex's robustness across various lookback configurations. As detailed in Table~\ref{tab:apex_lookback}, the default setting of a lookback window of 5 achieves a score of 52.3. Narrowing the window to 3 restricts the historical context, yielding a lower score of 50.3. Conversely, expanding the window to 10 degrades performance slightly to 50.6, as the broader scope begins to reintroduce the outdated, stale signals that our dynamic frontier strategy aims to eliminate. This confirms that a moderately sized lookback window provides the ideal balance between context retention and recency bias.

\begin{table}[h]
    \centering
    \caption{Sensitivity of the lookback window in \apex on IFBench with Gemini 2.5 Flash.}
    \label{tab:apex_lookback}
    \begin{tabular}{lc}
        \toprule
        \textbf{Setting} & \textbf{Score} \\
        \midrule
        lookback = 3 & 50.3 \\
        lookback = 5 (default) & 52.3 \\
        lookback = 10 & 50.6 \\
        \bottomrule
    \end{tabular}
\end{table}

\begin{figure}[h]
    \centering
    \includegraphics[width=0.5\linewidth]{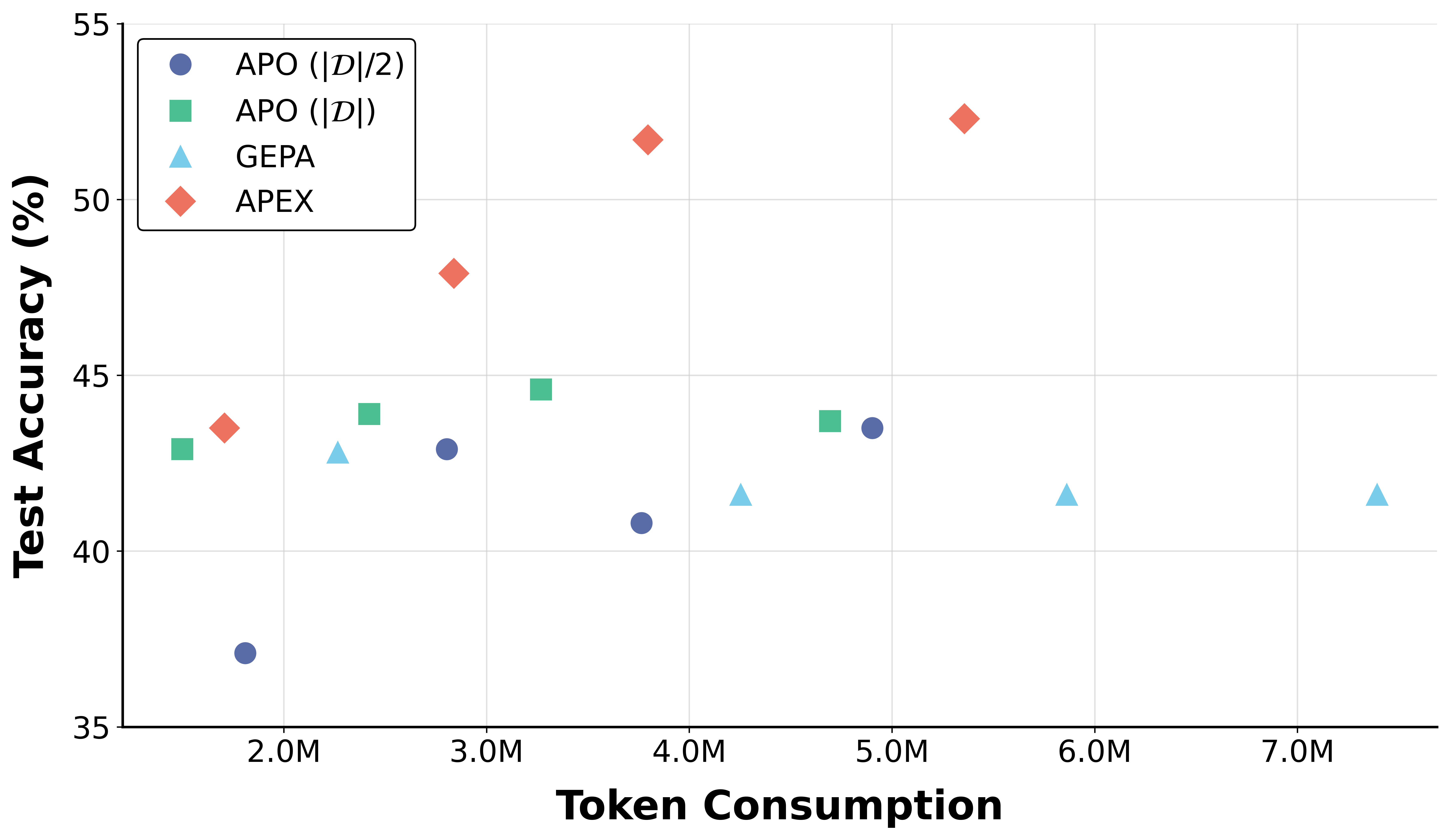}
    \caption{Test accuracy versus token consumption on IFBench with Gemini 2.5 Flash.}
    \label{fig:token}
\end{figure}

\paragraph{Token Cost.}
Figure \ref{fig:token} illustrates the trade-off between test accuracy and token consumption for API calls. The results demonstrate that APEX establishes a clear Pareto frontier, achieving higher test accuracy than the baseline methods.

\paragraph{Comparison of Prompt Length.}
Table~\ref{tab:prompt_length} demonstrates that \apex's superior performance is not a byproduct of generating longer prompts. While baseline methods like GEPA tend to continuously accumulate instructions, resulting in highly inflated prompt lengths, \apex achieves the highest overall accuracy with the most concise prompt. This confirms that \apex does not derive its performance gains from simply introducing more text. Instead, its optimization process successfully distills the instructions to their most critical and precise components, providing highly effective model guidance without redundant bloat.

\begin{table}[htbp]
    \centering
    \caption{Prompt lengths optimized by different methods on IFBench with Gemini 2.5 Flash.}
    \label{tab:prompt_length}
    \begin{tabular}{lc}
        \toprule
        \textbf{Method} & \textbf{Prompt Length} \\
        \midrule
        \apex & 257 \\
        APO ($|D|$) & 275 \\
        APO ($|D|/2$) & 342 \\
        GEPA & 662 \\
        \bottomrule
    \end{tabular}
\end{table}

\section{Initial Prompts}
\begin{lstlisting}[
    caption={Initial Prompt for IFBench}, 
    label={lst:mutation_prompt},
    basicstyle=\ttfamily\scriptsize, 
    lineskip=-0.5pt,
    breaklines=true,
    frame=single,
    backgroundcolor=\color{gray!5},
    captionpos=b
]

Respond to the given query. Enclose the final response in <answer> tags to distinguish it from the rest of your output.

Query: {query}
\end{lstlisting}

\begin{lstlisting}[
    caption={Initial Prompt for SimpleQA Verified}, 
    label={lst:mutation_prompt},
    basicstyle=\ttfamily\scriptsize, 
    lineskip=-0.5pt,
    breaklines=true,
    frame=single,
    backgroundcolor=\color{gray!5},
    captionpos=b
]

Respond to the given query. Your final answer must be in <answer> tags.

Query: {query}
\end{lstlisting}

\begin{lstlisting}[
    caption={Initial Prompt for FACTS Grounding}, 
    label={lst:mutation_prompt},
    basicstyle=\ttfamily\scriptsize, 
    lineskip=-0.5pt,
    breaklines=true,
    frame=single,
    backgroundcolor=\color{gray!5},
    captionpos=b
]

Answer the question based on the given context.
{user_request}
{context_document}
\end{lstlisting}

\section{APEX Prompt Templates}
\begin{lstlisting}[
    caption={APEX Critique Prompt}, 
    label={lst:prompt},
    basicstyle=\ttfamily\scriptsize, 
    lineskip=-0.5pt,
    breaklines=true,
    frame=single,
    backgroundcolor=\color{gray!5},
]
You are an expert prompt engineer. Your sole function is to analyze a faulty prompt and recommend the single most impactful, generalizable change for the next optimization iteration. Adhere strictly to the following process.

### The Prompt Under Analysis
The prompt being evaluated.
<current_prompt>
{prompt}
</current_prompt>

---
### Failure Case Analysis
Specific examples where the prompt produced a suboptimal output.
<failure_cases>
{error_cases}
</failure_cases>

---
### Instructions

**Step 1: Diagnose the Root Cause**
Analyze the **Failure Case Analysis** to identify the primary, underlying reasons for most of the errors. Classify the root cause into one of two major types:

* **Type 1: Weak Decision Boundaries (Defining the "What")**
    * **Ambiguity & Vague Definitions:** Terms, tone, or success criteria are open to interpretation.
    * **Constraint Loopholes:** Missing exclusionary constraints allow unwanted behaviors.
    * **Input Confusion:** The prompt lacks delimiters (e.g., XML tags, quotes), causing the model to confuse user input with instructions.
    * **Missing Context/Grounding:** The prompt fails to explicitly bind the model to provided source material (leading to hallucinations).

* **Type 2: Missing Process Instructions (Defining the "How")**
    * **Cognitive Overload:** The task is too complex for a single instruction and requires explicit decomposition (breaking the problem into distinct sub-tasks).
    * **Implicit Logic:** The prompt assumes the model knows the specific algorithm required to transform input to output.

**Step 2: Formulate the Recommendation**
Based on your diagnosis in Step 1, construct the recommendation according to these principles:

* **1. Apply the Correct Remediation Strategy:**
    * *If Type 1 (Boundary Issues):*
        * **Operationalize Definitions:** Change subjective adjectives to objective metrics.
        * **Enforce Delimiters:** Recommend wrapping input data in explicit tags (e.g., <input>...</input>) to separate it from instructions. Note that the placeholders and the corresponding input content are non-editable.
        * **Strengthen Constraints:** Add negative constraints or "Grounding" instructions (e.g., "Answer only using the provided text").
    * *If Type 2 (Process Issues):*
        * **Request Rationale:** Require the model to "show its work" or think step-by-step before answering. It is allowed to generate intermediate outputs before the final answer if a clear output format is specified for answer extraction.
        * **Decompose the Task:** Break the prompt into sequential, modular steps or sub-prompts.
        * **Few-Shot Prompting:** If the logic is abstract, strictly require "Input -> Output" examples to demonstrate the pattern.

* **2. Provide Generalizable Principles:**
    * Your recommendation must address the root cause, not just the specific failure examples.
    * **Crucially, do not quote or directly reference the provided `<failure_cases>`.** Your feedback must be independent of the specific content of the examples.
    * If an example is absolutely essential to illustrate your point, **you must invent a new, concise, and clear one** that demonstrates the principle effectively.

**Step 3: Construct the Feedback Object**
Translate your diagnosis into a structured directive for the Editor. You must use these exact fields:
* **Locator:** Quote the exact text, section header, or placeholder in the `<current_prompt>` where the fix should be applied.
* **Diagnosis:** Explain the specific weakness identified in Step 1 (e.g., "Type 1 Ambiguity: The adjective 'short' is subjective.").
* **Instruction:** The specific action for the Editor to take (e.g., "Replace 'short' with 'maximum 50 words'.").

---
### Required Output Format

You may provide explanatory text or rationale first (e.g., "Analysis: ...").
Your final response must be enclosed in `<actionable_feedback>` tags.
Inside these tags, you must strictly follow this format:

<actionable_feedback>
**Locator:** [Quote text or Header in the `<current_prompt>`]
**Diagnosis:** [Brief Type 1/Type 2 explanation]
**Instruction:** [Precise editing instruction]
</actionable_feedback>
\end{lstlisting}

\begin{lstlisting}[
    caption={APEX Mutation Prompt}, 
    label={lst:mutation_prompt},
    basicstyle=\ttfamily\scriptsize, 
    lineskip=-0.5pt,
    breaklines=true,
    frame=single,
    backgroundcolor=\color{gray!5},
    captionpos=b
]
You are an Adaptive Prompt Editor. Your goal is to rewrite a prompt based on targeted structured feedback.

### Input Data
**1. Original Prompt:**
<current_prompt>
{prompt}
</current_prompt>

**2. Critical Feedback:**
<feedback>
{feedback}
</feedback>
*(This feedback contains a Locator, a Diagnosis, and an Instruction.)*

---
### Execution Protocol

**1. Analysis Phase**
Before rewriting, explicitly plan your edit based on the input structure:
* **Target:** Find the specific text cited in the feedback's **Locator**.
* **Context:** Read the **Diagnosis** to understand the intent (this ensures you don't fix the grammar but miss the point).
* **Constraint Verification:** Ensure your planned edit does not accidentally remove critical constraints, negative instructions, or variable placeholders (e.g., `{{variable}}`).

**2. Revision Phase**
Rewrite the prompt using the following **"Logic vs. Syntax"** rules:
* **Variable Lockdown (Strict):** Treat all placeholders (e.g., `{{variable}}`) as immutable constants. Do not introduce any new placeholders, and do not add, remove, rename, or reformat existing ones. The content represented by the placeholder is non-editable.
* **Logic Lock (Strict):** Do not remove or alter instructions, constraints, or steps that are *not* targeted by the **Locator**.
* **Contextual Integration (Flexible):** You *are* permitted to adjust the wording, transitions, and grammar of the sentences surrounding the **Locator** to ensure the new changes blend naturally. The final result should read as a unified document.

---
### Output Format

**Part 1: Edit Strategy**
Provide a comprehensive, step-by-step plan. You must include:
* **The specific text/section you will modify.**
* **How you will rephrase it to satisfy the Instruction.**
* **How you will ensure surrounding transitions remain smooth.**
* **Confirmation that specific variables/constraints are preserved.**

**Part 2: Revised Prompt**
Output the full text of the revised prompt strictly enclosed within `<new_instruction>` tags.
\end{lstlisting}

\begin{lstlisting}[
    caption={APEX Error Case Template}, 
    label={lst:failure_example},
    basicstyle=\ttfamily\scriptsize, 
    lineskip=-0.5pt,
    breaklines=true,
    frame=single,
    backgroundcolor=\color{gray!5},
    captionpos=b
]
### Failure Example (Score: {score})

**1. Input Context:**
<input>
{query}
</input>

**2. Actual Model Output:**
<actual_output>
{response}
</actual_output>

**3. Evaluation Feedback (Why this failed):**
<critique>
{feedback}
</critique>
\end{lstlisting}

\section{Limitation}

While \apex improves the data efficiency of prompt optimization, it shares some fundamental limitations inherent to automated prompt engineering. First, the framework heavily relies on the availability of a representative dataset and a reliable, programmatic evaluation function, which can be challenging to obtain for highly subjective or open-ended tasks. Second, \apex's effectiveness is bounded by the target model's underlying capacities. It assumes the model possesses a reasonable baseline capability to perform the task when properly instructed, and it cannot overcome a fundamental lack of model knowledge or reasoning. Finally, our current evaluation primarily focuses on text-based LLMs. Future work could extend \apex to more diverse and complex scenarios, such as agentic workflows and multimodal tasks.

\section{Future Work}

The \apex framework opens up several promising dimensions for subsequent research. First, \apex can serve as a versatile, plug-and-play core optimization layer within established LLM programming frameworks~\citep{yuksekgonul2025optimizing, khattab2024dspy}. Second, \apex's underlying mechanism can be leveraged to improve inference-time scaling techniques~\citep{novikov2025alphaevolve, snell2025scaling} by dynamically guiding test-time compute allocation and refining intermediate search paths. Finally, future work can extend \apex beyond static, text-based scenarios to multimodal tasks, complex agentic workflows, and highly interactive environments. For example, \apex could be utilized to optimize contextual text rephrasing to shape intricate downstream behaviors, such as prompting knowledge sharing via causal language in social networks~\citep{shi2024diffusion}.

More broadly, our findings highlight a fundamental, yet underexplored principle for black-box LLM optimization. As candidate solutions evolve, the target data distribution dynamically co-evolves. Successfully modeling, tracking, and stabilizing this shifting landscape represents a crucial frontier for the robust optimization of black-box LLMs.

\section{Impact Statement}

This work significantly improves accessibility to high-performance LLMs by automating complex prompt engineering. By removing the bottleneck of manual expertise, \apex allows a broader audience, including non-experts and smaller organizations, to achieve state-of-the-art results previously reserved for specialized teams. Furthermore, this efficiency drives a critical shift in computational resource allocation. By demonstrating that the performance gap between optimized prompts and expensive parameter updates is rapidly narrowing, \apex encourages a move towards more sustainable, energy-efficient model adaptation strategies, fundamentally reducing the computational overhead of AI deployment.
However, we acknowledge that automated prompt optimization is a dual-use capability. The same mechanisms that efficiently navigate the prompt space could be misused. \apex operates as an orchestration layer built on top of LLMs, which serve as the final execution engines. As LLMs continue to grow safer, they inherently ensure that optimization tools like \apex remain securely bounded, regardless of how a prompt is engineered.

\end{document}